\documentclass[conference]{IEEEtran}
\IEEEoverridecommandlockouts
\usepackage{cite}
\usepackage{amsmath,amssymb,amsfonts}
\usepackage{algorithmic}
\usepackage[ruled, vlined, linesnumbered]{algorithm2e}
\usepackage{graphicx}
\usepackage{subcaption}
\usepackage{textcomp}
\usepackage{xcolor}
\usepackage{bm}
\usepackage{bbm}
\usepackage{booktabs}
\usepackage{microtype}
\usepackage{multirow}
\usepackage{cleveref}
\usepackage{xspace}
\usepackage{xcolor}
\usepackage{enumitem}
\usepackage{placeins}
\usepackage{soul}

\newcommand{\huimin}[1]{\textcolor{black}{#1}}

\begin{document}

\title{Unsupervised Domain Adaptation for COVID-19 Information Service with Contrastive Adversarial Domain Mixup}

\author{\IEEEauthorblockN{Huimin Zeng\IEEEauthorrefmark{1}, Zhenrui Yue\IEEEauthorrefmark{1}, Ziyi Kou\IEEEauthorrefmark{1}, Lanyu Shang\IEEEauthorrefmark{1}, Yang Zhang\IEEEauthorrefmark{2}, Dong Wang\IEEEauthorrefmark{1}}
\IEEEauthorblockA{\IEEEauthorrefmark{1}School of Information Sciences\\
University of Illinois Urbana-Champaign, IL, USA\\
\{huiminz3, zhenrui3, ziyikou2, lshang3, dwang24\}@illinois.edu}
\IEEEauthorblockA{\IEEEauthorrefmark{2}Department of Computer Science and Engineering\\
University of Notre Dame, IN, USA\\
yzhang42@nd.edu}}

\maketitle

\begin{abstract}
In the real-world application of COVID-19 misinformation detection, a fundamental challenge is the lack of the labeled COVID data to enable supervised end-to-end training of the models, especially at the early stage of the pandemic. To address this challenge, we propose an unsupervised domain adaptation framework using contrastive learning and adversarial domain mixup to transfer the knowledge from an existing source data domain to the target COVID-19 data domain. In particular, to bridge the gap between the source domain and the target domain, our method reduces a radial basis function (RBF) based discrepancy between these two domains. Moreover, we leverage the power of domain adversarial examples to establish an intermediate domain mixup, where the latent representations of the input text from both domains could be mixed during the training process. Extensive experiments on multiple real-world datasets suggest that our method can effectively adapt misinformation detection systems to the unseen COVID-19 target domain with significant improvements compared to the state-of-the-art baselines.
\end{abstract}


\section{Introduction}
\label{sec:intro}
\huimin{In this work, we focus on COVID-19 misinformation detection, given its global impact of the ongoing pandemic and the ``Infodemic''\footnote{https://www.who.int/health-topics/infodemic\#tab=tab\_1} it causes on social media \cite{malla2021covid}}. Regarding COVID-19 misinformation detection, if the language models trained on non-COVID datasets without any fine-tuning on COVID-19 specific data, these models might suffer from a severe issue of generalization and perform poorly on the COVID-19 data, due to the domain shift between the non-COVID training data distribution and the test COVID-19 data distribution. Recently, the ongoing pandemic of COVID-19 inspires a variety of studies \cite{kou2022crowd} to develop NLP models to provide reliable COVID-19 information services across various social media platforms (e.g., Twitter, Facebook). However, the supervised learning approaches often require a large-scale training dataset while collecting annotations for COVID training data is extremely expensive and time consuming due to the cost and complexity in recruiting the qualified annotators and keep the annotations update to date to accommodate the dynamics of COVID-19 knowledge (e.g., different variants of the virus) \cite{oniani2020qualitative}. Moreover, our unsupervised domain adaptation setting is motivated for a more general setting of any early-stage pandemic (not limited to COVID-19) where there is no ground-truth information about the novel disease at all, but the need for correct information is urgent. Therefore, it is critical to develop unsupervised domain adaptation frameworks to train COVID models so that knowledge from an existing data domain could be adapted and transferred to the unseen COVID data domain without requiring any ground-truth training labels.

\begin{figure*}[!thb]
\centering
\begin{subfigure}{0.31\linewidth}
  \includegraphics[width=\linewidth]{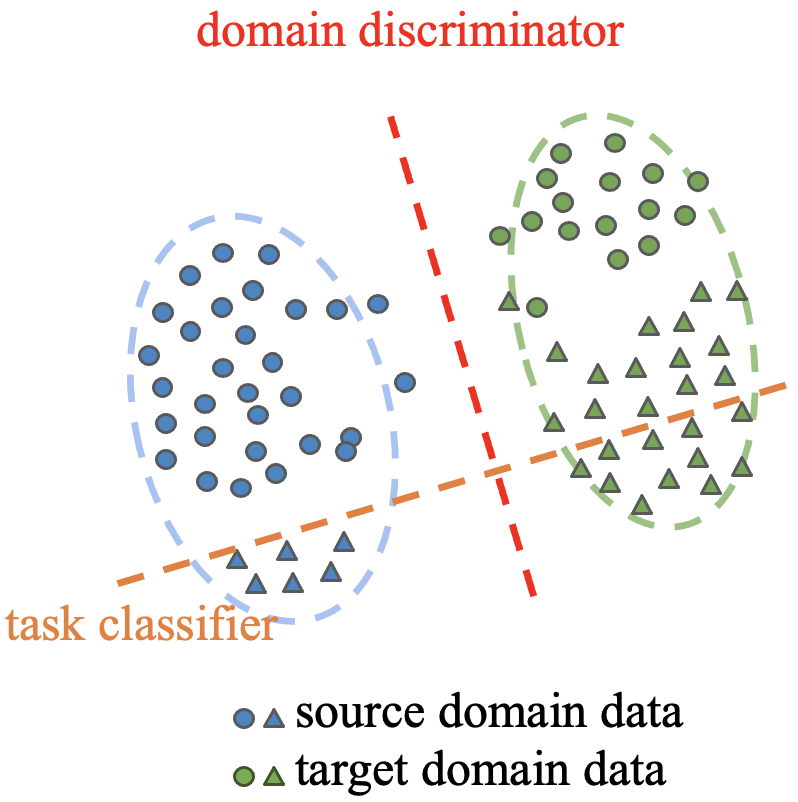}
  \caption{Unadapted}
  \label{fig:intuition_0}
\end{subfigure}
\begin{subfigure}{0.32\linewidth}
  \centering
  \includegraphics[width=\linewidth]{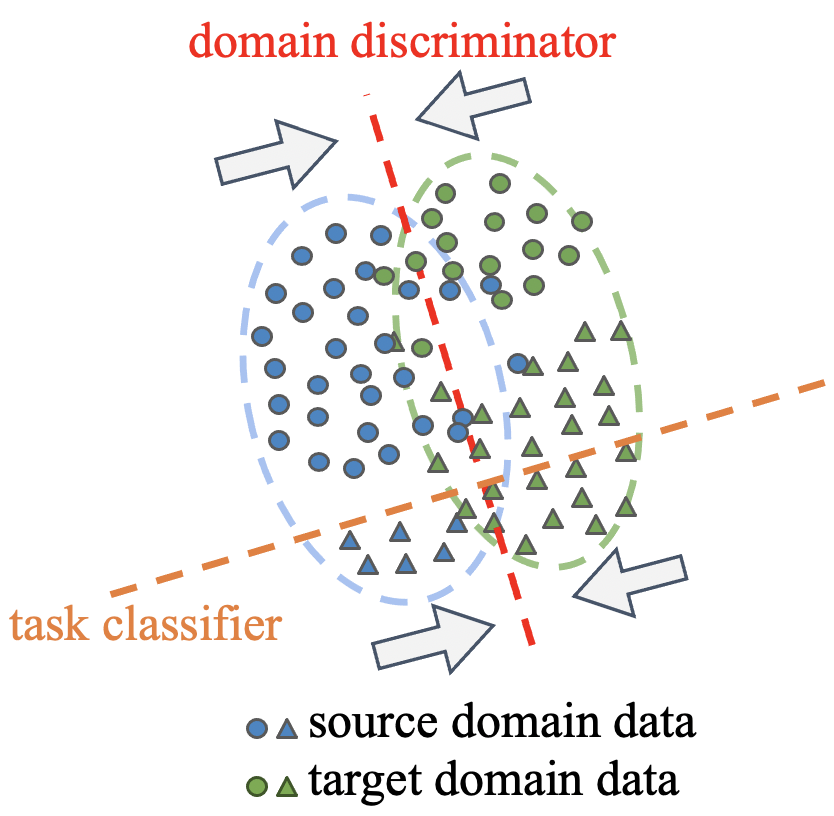}
  \caption{Adversarial Domain Mixup}
  \label{fig:intuition_1}
\end{subfigure}
\begin{subfigure}{0.31\linewidth}
  \centering
  \includegraphics[width=\linewidth]{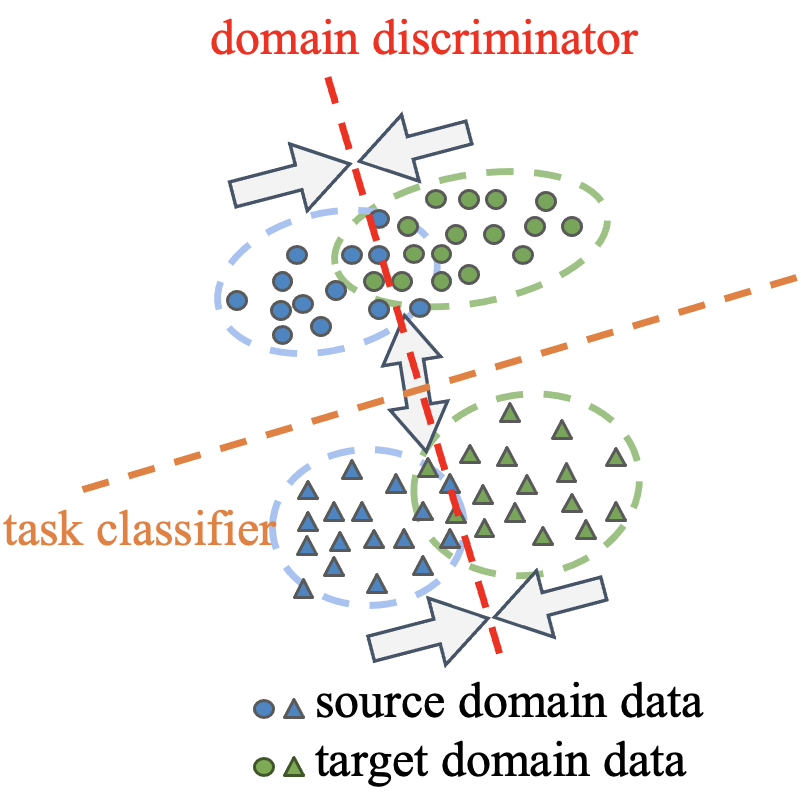}
  \caption{Constrastive Domain Adaptation}
  \label{fig:intuition_2}
\end{subfigure}
\caption{\huimin{The Overview of Our Contrastive Adversarial Domain Mixup (CADM):} firstly, in (a), a pre-trained models will generate labels for target domain examples via pseudo labeling, where the green triangles belong to one class (e.g., true information in the misinformation detection task) and the green dots belong to another class (e.g., false information in the misinformation detection task). In addition, a domain discriminator is trained. Then, in (b), the well-trained domain discriminator will establish an intermediate domain mixup by perturbing latent representations of input text from both domains towards each other. At the same time, in (c), we also compute a contrastive adaptation loss over the perturbed adversarial representations, and optimize it to further reduce the domain discrepancy and increase the models' performance. Note that (b) and (c) are executed alternately.}
\label{fig:intuition}
\end{figure*}

In this paper, we explore an unsupervised domain adaptation problem for COVID-19 misinformation detection on social media. In particular, we propose an unsupervised domain adaptation framework \huimin{\textbf{C}ontrastive \textbf{A}dversarial \textbf{D}omain \textbf{M}ixup (CADM)}, which uses adversarial domain mixup and contrastive learning to bridge the gap between the source training data domain and the target COVID data domain. The overview of our framework is shown in Figure~\ref{fig:intuition}. To demonstrate the effectiveness of the proposed CADM, we evaluate it \huimin{on several real-world COVID-19 datasets}. Our experimental results suggest that our CADM effectively adapts pre-trained language models to the target COVID domain, and consistently outperforms state-of-the-art baselines.



    

    
    
\section{Related Work}
\label{sec:related_work}

\textbf{Misinformation Detection.} Great efforts have been made to detect the misinformation from online platforms (e.g., social media). In \cite{kou2022crowd}, knowledge graphs are integrated into the misinformation detection framework to enhance the model's performance. The concurrent work \cite{yue2022contrastive} also proposed a domain adaptation framework to address the COVID-19 misinformation detection using label correction. However, such misinformation detection systems are built under a supervised or semi-supervised learning setting, but in practice labeled COVID-19 misinformation data is not always accessible. Therefore, this paper focuses on \textit{unsupervised} domain adaptation of language models for COVID-19 misinformation detection, where the models are trained to adapt knowledge from a source 
domain to the unknown COVID-19 domain.

\textbf{Domain Adaptation.}
\huimin{Only limited domain adaptation} methods are developed for misinformation detection. In \cite{zou2021unsupervised}, domain adversarial training is implemented for misinformation task, so that the models are trained to learn domain-invariant features. Utilizing contrastive methods, \cite{yue2021contrastive} propose to quantify and reduce the domain discrepancy using explicit distance measures (e.g., maximum mean discrepancy) to bridge the gap between source domain and target domain. \huimin{However, the unsupervised domain adaptation methods have not been systematically studied in the COVID-19 domain.} In this work, inspired by the idea of adversarial examples \cite{zeng2021adversarial,madry2017towards} and domain mixup \cite{xu2020adversarial}, we propose to establish a smoothed intermediate training domain by perturbing the latent representations of the input from both source domain and target domain towards each other with a domain discriminator and perform contrastive training on the smoothed domain to transfer knowledge from the source training domain to the target COVID-19 domain.

\section{Problem Statement}
\label{sec:problem}
Regarding misinformation detection, we aim at training a model $f$, which takes an input text $\bm{x}$ (a COVID-19 claim or a piece of news) to predict whether the information contained in $\bm{x}$ is valid or not (i.e., a binary classification task). Moreover, in our domain adaptation problem, we use $\mathcal{P}$ to denote source domain data distribution and $\mathcal{Q}$ for the target domain data distribution. Each data point $(\bm{x}, y)$ contains an input segment of COVID-19 claim or news ($\bm{x}$) and a label $y \in \{0,1\}$ ($y=1$ for true information and $y=0$ for false information). To differentiate the notations of the data sampled from the source distribution $\mathcal{P}$ and the target distribution $\mathcal{Q}$, we further introduce two \huimin{definitions of the domain data}:

\begin{itemize}
  \item \emph{\textbf{Source domain}}: The subscript $s$ is used to denote the source domain data: $\bm{\mathcal{X}}_s = \{(\bm{x}_s, y_s)|(\bm{x}_s, y_s)\sim\mathcal{P}\}$.
  
   \item \emph{\textbf{Target domain}}: Similarly, the subscript $t$ is used to denote the target domain data: $\bm{\mathcal{X}}_t = \{\bm{x}_t|\bm{x}_t\sim\mathcal{P}\}$. Note that in our unsupervised setting, the ground truth labels \huimin{of target domain data} $y_t$ are not used during training.
\end{itemize}

Our goal is to adapt a classifier $f$ trained on $\mathcal{P}$ to $\mathcal{Q}$. For a given target domain input $\bm{x}_t$, a well-adapted model aims at making predictions as correctly as possible. 


\section{Solution}
\label{sec:alg}
\textbf{Domain Discriminator.} The first step of our framework is to train a domain discriminator $f_D$ to classify whether the input data belongs to the source or target domain. The domain discriminator shares the same BERT Encoder $\bm{f}_e$ with the COVID model and has a different binary classification module $f_D$. The domain discriminator takes the token \texttt{[CLS]} representation from the BERT encoder as input to predict the domain of the input data as in \cite{lee2019domain}:
\begin{equation}
    \hat{y} = f_D(\bm{z}),
\end{equation}
where $\bm{z}$ is the representation of token \texttt{[CLS]}.
    
Regarding the training of the $f_D$, we explicitly define the domain label $y_D$ of the source domain data as $y_D=0$ and the domain label of the target domain data as $y_D=1$. Note that the domain labels $y_D$ are not the same as data label $y$ (true information or false information). Therefore, the training of the domain discriminator is formulated as:
\begin{equation}
\label{eq:domain_train_mis}
  \min_{f_D}\mathbb{E}_{(\bm{x}, y_D) \sim \mathcal{X}'}
\Big[l(f_{D}(f_e(\bm{x})), y_D)\Big],
\vspace{-0.1cm}
\end{equation}
where \huimin{$\mathcal{X}'$} represents the merged datasets of both source domain and target domain training data with domain labels.

\textbf{Adversarial Domain Mixup.} After the domain discriminator is trained, we propose to directly perturb the latent representations of the input data from both source domain and target domain towards the decision boundary of the domain discriminator \huimin{as shown in Figure~\ref{fig:intuition_1}}. To this end, the perturbed representations (i.e., domain adversarial representations) from both domains could become closer to each other, indicating a reduced domain gap. Herein, the generated domain adversarial representations from both domains form a smoothed intermediate domain mixup in the latent feature space of the model. Mathematically, the optimal perturbation $\bm{\delta}^*$ to perturb the latent representation $\bm{z}$ of a training sample $\bm{x}$ could be found by solving an optimization problem:
\vspace{-0.1cm}
\begin{equation}
\label{eq:adv_optimization}
\begin{split}
    \mathcal{A}(f_e, f_D, \bm{x}, y_D, \epsilon) = & \max_{\bm{\delta}} \Big[l(f_{D}(\bm{z} + \bm{\delta}), y_D)\Big] \\
    & s.t. \quad \| \bm{\delta} \| \leq \epsilon, \quad \bm{z} = f_e(\bm{x}).
\end{split}
\vspace{-0.15cm}
\end{equation}
Note that in the above equation, we introduce a hyperparameter $\epsilon$ to bound the norm of the perturbation $\delta$, so that the infinity solution could be avoided. Eventually, after applying Equation~\ref{eq:adv_optimization} to all training samples in the merged training set $\mathcal{X}'$, we obtain the adversarial domain mixup $\mathcal{Z}'$:
\vspace{-0.1cm}
\begin{equation}
\label{eq:adv_domain_mixup}
\begin{split}
    \mathcal{Z}' &= \{\bm{z}'| \bm{z}'=\bm{z}+\mathcal{A}(f_e, f_D, \bm{x}, y_D, \epsilon), (\bm{x},y_D) \in\mathcal{X}'\} \\ 
    & := \mathcal{Z}'_{s} \cup \mathcal{Z}'_{t},
\end{split}
\end{equation}
where $\mathcal{Z}'_{s}$ are perturbed source features and $\mathcal{Z}'_{t}$ are perturbed target features. We use the projected gradient descent (PGD) to approximate the solution of Equation~\ref{eq:adv_optimization} as in \cite{madry2017towards,zeng2021adversarial}.

\textbf{Contrastive Domain Adaptation.} Next, inspired by \cite{yue2021contrastive}, we propose a two-fold contrastive adaptation loss over $\mathcal{Z}_{adv}$ to further adapt knowledge from the source data domain to the target data domain. Firstly, we reduce the domain discrepancy among intra-class representations. That is, if a representation from the source data domain has a label of being true \huimin{(or false)} and a representation from the target data domain has a pseudo label of being true \huimin{(or false)}, then these two representations are considered as intra-class representations and we reduce the domain discrepancy between them. The second level of our contrastive adaptation loss is defined for inter-class representations. \huimin{As shown in Figure~\ref{fig:intuition_2}}, the discrepancy between the representations of true information and false information will be enlarged. 

To compute the our proposed contrastive adaptation loss, we propose to measure the discrepancy among token classes using radial basis functions (RBF). In \cite{van2020uncertainty}, RBF is proved to be an efficient tool to quantify uncertainty in deep neural networks. Since our pseudo labeling process \huimin{is designed to} automatically filter out low-confident labels for the target domain data, using RBF to measure the discrepancy among token classes could efficiently improve the quality of the pseudo labels and ultimately contribute the domain adaptation of the model. Formally, with the definition of the RBF kernel $k(z_1,z_2)=\mathrm{exp}[-\frac{\|\bm{z}_1 - \bm{z}_2\|^2}{2\sigma^2}]$, we define the class-aware loss for the misinformation detection task as follows:
\vspace{-0.15cm}
\begin{equation}
\label{eq:l_rbf_mis}
    \begin{split}
        \mathcal{L}_{con}(\bm{\mathcal{Z}}') &= - \sum_{i=1}^{|\bm{\mathcal{Z}}'_{s}|} \sum_{j=1}^{|\bm{\mathcal{Z}}'_{t}|} \frac{\mathbbm{1}(y_s^{(i)}=0, \hat{y}_t^{(j)}=0)k(\bm{z}_s^{(i)}, \bm{z}_t^{(j)})}{\sum_{l=1}^{|\bm{\mathcal{Z}}'_{s}|} \sum_{m=1}^{|\bm{\mathcal{Z}}'_{t}|} \mathbbm{1}(y_s^{(l)}=0, \hat{y}_t^{(m)}=0)} \\
        & - \sum_{i=1}^{|\bm{\mathcal{Z}}'_{s}|} \sum_{j=1}^{|\bm{\mathcal{Z}}'_{t}|} \frac{\mathbbm{1}(y_s^{(i)}=1, \hat{y}_t^{(j)}=1)k(\bm{z}_s^{(i)}, \bm{z}_t^{(j)})}{\sum_{l=1}^{|\bm{\mathcal{Z}}'_{s}|} \sum_{m=1}^{|\bm{\mathcal{Z}}'_{t}|} \mathbbm{1}(y_s^{(l)}=1,\hat{y}_t^{(m)}=1)} \\
         & + \sum_{i=1}^{|\bm{\mathcal{Z}}'_{s}|} \sum_{j=1}^{|\bm{\mathcal{Z}}'_{s}|} \frac{\mathbbm{1}(y_s^{(i)}=1, y_s^{(j)}=0)k(\bm{z}_s^{(i)}, \bm{z}_s^{(j)})}{\sum_{l=1}^{|\bm{\mathcal{Z}}'_{s}|} \sum_{m=1}^{|\bm{\mathcal{Z}}'_{s}|} \mathbbm{1}(y_s^{(l)}=1,y_s^{(m)}=0)} \\
          & + \sum_{i=1}^{|\bm{\mathcal{Z}}'_{t}|} \sum_{j=1}^{|\bm{\mathcal{Z}}'_{t}|} \frac{\mathbbm{1}(\hat{y}_t^{(i)}=1, \hat{y}_t^{(j)}=0)k(\bm{z}_t^{(i)}, \bm{z}_t^{(j)})}{\sum_{l=1}^{|\bm{\mathcal{Z}}'_{t}|} \sum_{m=1}^{|\bm{\mathcal{Z}}'_{t}|} \mathbbm{1}(\hat{y}_t^{(l)}=1,\hat{y}_t^{(m)}=0)},
    \end{split}
\vspace{-0.15cm}
\end{equation}
where $\hat{y}_t$ is the pseudo label of the target domain samples and $\bm{z}$ denotes the representation of token \texttt{CLS}.

\textbf{Overall Contrastive Adaptation Loss.} Now, we merge the cross-entropy loss of the task classification problem and the above contrastive adaptation loss into a single optimization objective for the COVID model:
\vspace{-0.1cm}
\begin{equation}
\label{eq:loss_all}
    \mathcal{L}_{all} = \mathcal{L}_{ce}(\bm{\mathcal{X}}) + \lambda\mathcal{L}_{con}(\bm{\mathcal{Z}}'),
\vspace{-0.1cm}
\end{equation}
where $\mathcal{L}_{ce}$ is the cross-entropy loss over the training data with ground-truth label or the pseudo label, and $\lambda$ is the hyperparameter to adjust the domain adaptation strength. Moreover, we compute the RBF kernel with multiple bandwidths for $\mathcal{L}_{con}$, since multiple bandwidths of the RBF kernel encourage the model to learn a smoothed and generalized feature space \cite{yue2021contrastive}. 

\section{Evaluation}
\label{sec:exp}
\subsection{Experimental Design}
In our experiments, we use three source misinformation datasets (GossipCop~\cite{shu2020fakenewsnet}, LIAR~\cite{wang2017liar} and PHEME~\cite{buntain2017automatically}) released \textit{before} the COVID outbreak and two COVID misinformation datasets (Constraint~\cite{patwa2021fighting} and ANTiVax~\cite{hayawi2022anti}) collected \textit{after} the outbreak as target datasets. Following~\cite{zou2021unsupervised}, the commonly used RoBERTa \cite{liu2019roberta} was selected as the misinformation detection model. Moreover, regarding the baseline methods, we select DAAT \cite{du2020adversarial}, where the misinformation detection model is post-trained to improve the domain-adversarial adaptation, and EADA \cite{zou2021unsupervised}, where energy-based domain adversarial training is performed using autoencoder. Finally, in terms of the evaluation metrics, we use the balanced accuracy, accuracy and F1 score to evaluate the models' performance as in \cite{yue2022contrastive}.

\begin{table*}[!bth]
\centering
\caption{Results of domain adaptation for COVID-19 misinformation detection.}
\label{tab:mis_results}
\begin{tabular}{@{}lcccccccccc@{}}
\toprule
\multirow{2}{*}{\textbf{Target Dataset}} & \textbf{Source Dataset}          & \multicolumn{3}{c}{\textbf{LIAR}}      & \multicolumn{3}{c}{\textbf{GossipCop}}     & \multicolumn{3}{c}{\textbf{PHEME}}        \\ \cmidrule(l){2-11} 
                                         & Metric               & BA $\uparrow$   & Acc. $\uparrow$ & F1 $\uparrow$   & BA $\uparrow$   & Acc. $\uparrow$ & F1 $\uparrow$ & BA $\uparrow$   & Acc. $\uparrow$ & F1 $\uparrow$   
                                         \\ \midrule
\multirow{4}{*}{Constraint}              & No Adaptation                 & 0.7231    & 0.7322    & 0.7822    & 0.5638    & 0.5832    & 0.7110    & 0.4889    & 0.5047    & 0.6360 \\
                                         & DAAT                 & 0.7606    & 0.7626    & 0.7795    & 0.7178    & 0.7276    & 0.7806    & 0.5227    & 0.5411    & 0.6763 \\
                                         & EADA                 & 0.7776    & 0.7794    & 0.7950    & 0.5210    & 0.5430    & 0.6944    & 0.4944    & 0.4969    & 0.6391 \\
                                         & CADM (Ours)                 & \textbf{0.7858}   & \textbf{0.7944}    & \textbf{0.8317}  & \textbf{0.7787} & \textbf{0.7780} &  \textbf{0.7828} & \textbf{0.6430} & \textbf{0.6547} &  \textbf{0.7301}  
                                         \\ \midrule
\multirow{4}{*}{ANTiVax}                 & No Adaptation                 & 0.5444    & 0.4929    & 0.4162    & 0.5695    & 0.6501    & 0.7673    & 0.5294    & 0.6196    & 0.7531 \\
                                         & DAAT                 & 0.6228    & 0.5778    & 0.5393    & 0.6692    & 0.7161    & 0.7918    & 0.5895    & 0.6498    & 0.7518 \\
                                         & EADA                 & 0.6184    & 0.6642    & 0.7490    & 0.5509    & 0.6434    & 0.7709    & 0.5411    & 0.6328    & 0.7632 \\
                                         & CADM (Ours)                 & \textbf{0.7123}   & \textbf{0.7158}    & \textbf{0.7652}  & \textbf{0.7522} & \textbf{0.7701} &  \textbf{0.8152} & \textbf{0.6752} & \textbf{0.7323} &  \textbf{0.8107}  
                                         \\ \midrule
\bottomrule
\end{tabular}
\vspace{-0.2cm}
\end{table*}

\subsection{Evaluation Results}
The results are reported in Table~\ref{tab:mis_results}. We observe that our method consistently outperforms all baseline algorithms in terms of adapting the models to the unseen COVID-19 domain. For instance, consider the setting where the model is adapted \huimin{from} source domain PHEME to target domain Constraint. This setting is the most challenging one: the model can only make random guess on the target Constraint dataset (BA=0.4889) without any domain adaptation. Under this setting, EADA failed to adapt the model by achieving a balanced accuracy of 0.4944. As for the other baseline DAAT, the adapted model only performs slightly better, namely BA=0.5227. In contrast, the performance of the model trained using our framework could be significantly improved (e.g., BA=0.6430). Similar trends could be observed on other adaptation settings with different source and target domain combinations. We observe that our adapted models achieve a better performance than the baselines on all metrics. Moreover, we also observe that our CADM is more robust than the other two baseline methods. For instance, regardless of the source domain dataset and target domain dataset, our method could consistently adapt the misinformation detection model to the target domain. However, on some datasets, the baseline methods successfully adapt the model but on other datasets the baseline methods simply fail. For instance, from GossipCop to Constraint, the model was adapted towards a worse direction, when EADA is deployed. 

\section{Conclusion}
\label{sec:conclusion}
In this paper, we present a novel unsupervised domain adaptation framework for COVID-19 misinformation detection on social media data. Our unsupervised framework is motivated by the fact that the ground-truth labels of the COVID-19 data are not always available but the need for high-quality information services is always persistent and urgent. In addition to COVID-19, our method has the potential to provide efficient solutions to many other information services on social media platforms, when the training labels of target domain data are missing. Extensive experimental results suggest that our method could effectively adapt the misinformation detection models from source domain to the target COVID-19 domain without requiring labels of COVID-19 data.

\section*{Acknowledgments}

This research is supported in part by the National Science Foundation under Grant No. IIS-2202481, CHE-2105005, IIS-2008228, CNS-1845639, CNS-1831669. The views and conclusions contained in this document are those of the authors and should not be interpreted as representing the official policies, either expressed or implied, of the U.S. Government. The U.S. Government is authorized to reproduce and distribute reprints for Government purposes notwithstanding any copyright notation here on.

\bibliographystyle{IEEEtran}
\bibliography{reference}

\end{document}